\documentclass[10pt,twocolumn,letterpaper]{article}

\usepackage{iccv}
\usepackage{times}
\usepackage{epsfig}
\usepackage{graphicx}
\usepackage{amsmath}
\usepackage{amssymb}

\usepackage{multirow}
\usepackage{multicol}
\usepackage{times}
\usepackage{epsfig}
\usepackage{amsfonts}
\usepackage{array}
\usepackage{caption}
\usepackage{subcaption}
\usepackage{arydshln}
\usepackage{bm}
\usepackage{balance}
\usepackage{booktabs}
\usepackage{tikz}
\usepackage{pifont}
\usepackage{mathrsfs}
\newcommand{\cmark}{\ding{51}}

\newcommand{\dash}{\textendash}

\usepackage[pagebackref=true,breaklinks=true,letterpaper=true,colorlinks,bookmarks=false]{hyperref}

\iccvfinalcopy 

\def\iccvPaperID{11272} 

\ificcvfinal\fi

\begin{document}

\title{Vision Transformer with Attention Map Hallucination and FFN Compaction}
\newcommand*\samethanks[1][\value{footnote}]{\footnotemark[#1]}
\author{
Haiyang Xu$^{1,2}$\thanks{This work was done when H. Xu was intern at Baidu VIS, Beijing, P.R. China} \quad
Zhichao Zhou$^2$ \quad
Dongliang He$^2$ \quad
Fu Li$^2$ \quad
Jingdong Wang$^2$\thanks{Corresponding author.} \\[1.2mm]
$^1$University of Science and Technology of China \quad
$^2$Baidu VIS
}

\maketitle
\ificcvfinal\thispagestyle{empty}\fi

\begin{abstract}
Vision Transformer(ViT) is now dominating many vision tasks. The drawback of quadratic complexity of its token-wise multi-head self-attention (MHSA), is extensively addressed via either token sparsification or dimension reduction (in spatial or channel). However, the therein redundancy of MHSA is usually overlooked and so is the feed-forward network (FFN). 
To this end, we propose attention map hallucination and FFN compaction to fill in the blank. Specifically, we observe similar attention maps exist in vanilla ViT and propose to hallucinate half of the attention maps from the rest with much cheaper operations, which is called hallucinated-MHSA (hMHSA).
As for FFN, we factorize its hidden-to-output projection matrix and leverage the re-parameterization technique to strengthen its capability, making it compact-FFN (cFFN). With our proposed modules, a 10\%-20\% reduction of floating point operations (FLOPs) and parameters (Params) is achieved for various ViT-based backbones, including straight (DeiT), hybrid (NextViT) and hierarchical (PVT) structures, meanwhile, the performances are quite competitive.
\end{abstract}

\section{Introduction}
\label{sec:intro}

\begin{figure}
   \centering
    \begin{subfigure}[b]{\columnwidth}
        \centering
        \includegraphics[width=\columnwidth]{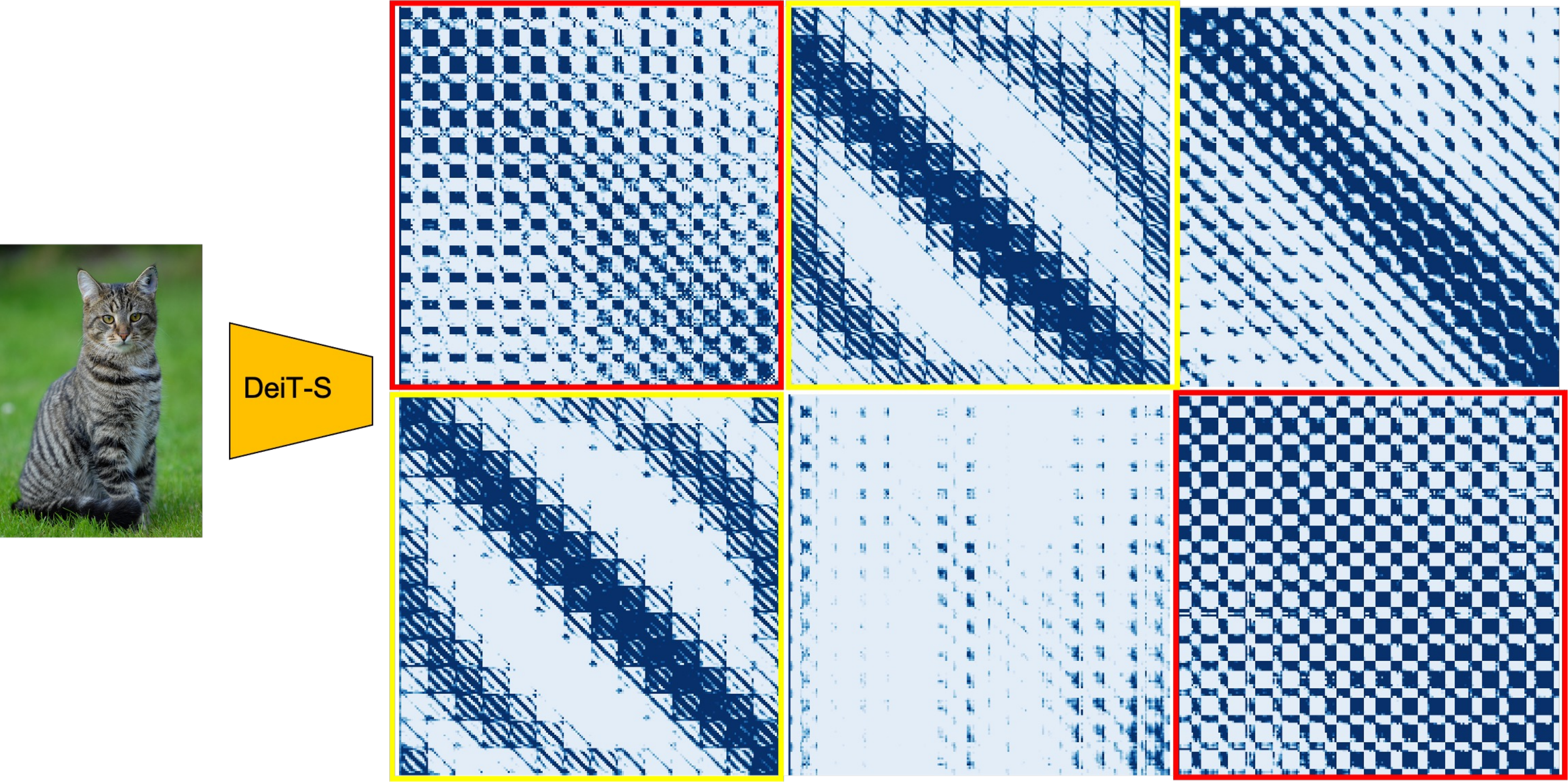}
        \caption{Visualization of the attention maps. }
        \label{fig:intro-q1}
    \end{subfigure}
    \begin{subfigure}[b]{\columnwidth}
        \centering
        \includegraphics[width=\columnwidth]{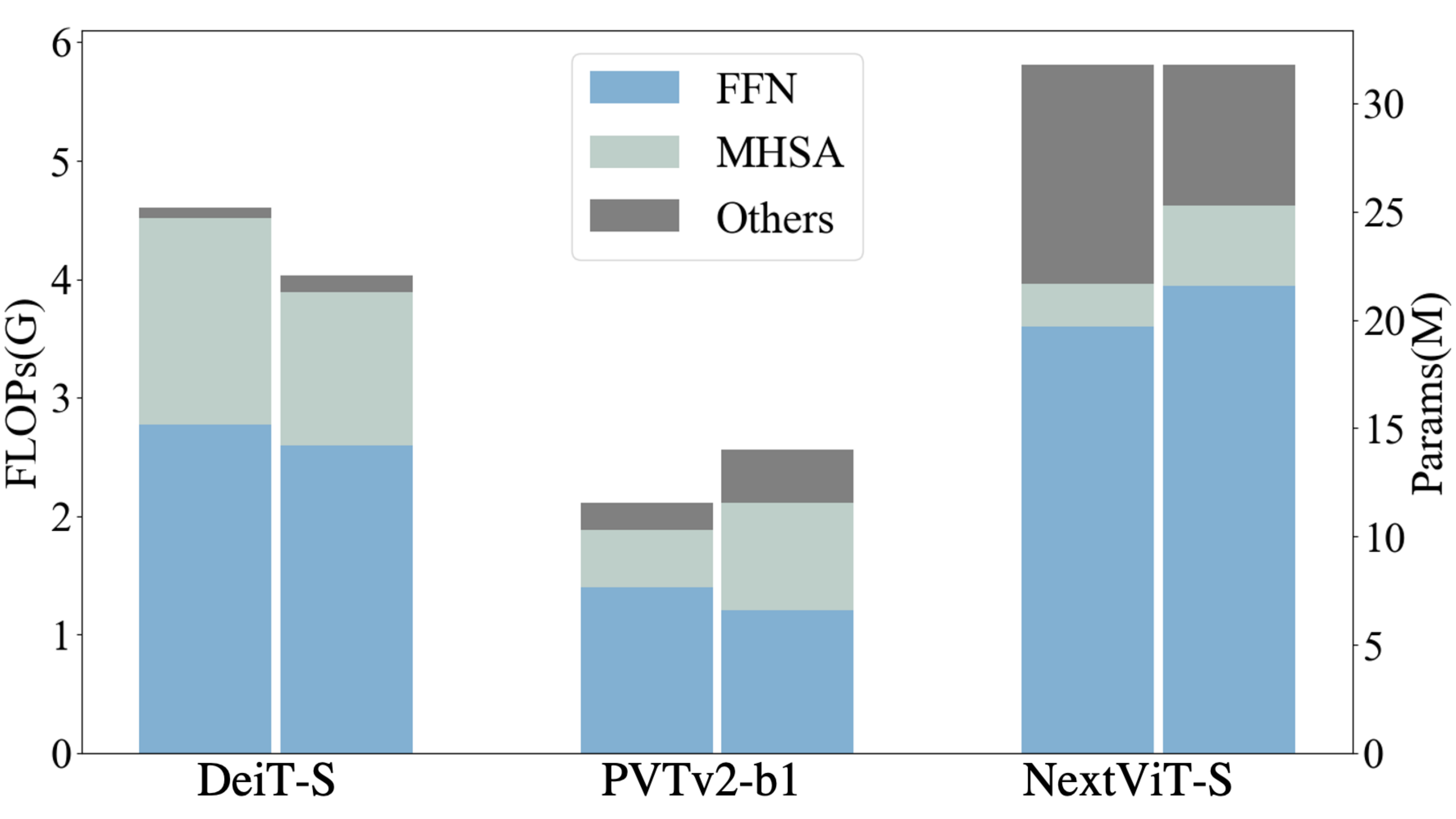}
        \caption{Statistics of model complexity.}
        \label{fig:intro-q2}
    \end{subfigure}
    \caption{(a) Attention maps of the $2^{nd}$ block of DeiT-S \cite{touvron2021deit} trained on ImageNet. It can be seen that there are two pairs of similar attention maps and we mark them with red and yellow borders, respectively. It suggests the expensive MHSA is redundant and correlated among its attention maps and can have better designs. (b) FLOPs (left) and Params (right) of three backbones\cite{touvron2021deit, wang2022pvtv2, li2022nextvit}. Complexity of MHSA, FFN and other modules are separately measured. Results show that the cost of FFN is nontrivial. Nevertheless, the cost is usually overlooked in vision transformer architecture design.}
\end{figure}

Recently, Vision Transformers (ViT)\cite{dosovitskiy2020vit, touvron2021deit, chu2021twinsvit, yuan2021t2tvit} have achieved significant performance improvement on various computer vision tasks, such as image classification\cite{liu2021swin, liu2022swinv2, dong2022cswin}, object detection\cite{carion2020DETR, li2022vitdet, zhang2022dino, zhu2020deformabledetr} and semantic segmentation\cite{zheng2021SETR, ranftl2021vitdensepred, strudel2021segmenter}. Meanwhile, it is widely agreed that vision transformers suffer from quadratic computational cost caused by the token-wise multi-head self-attention (MHSA) module. Such model complexity makes vision transformers inferior to convolutional neural networks (CNN)\cite{he2016deep, simonyan2014vgg, xie2017resnext, liu2022convnext} when inference speed is a critical factor in real-world applications, especially when compared to efficient CNN variants\cite{sandler2018mobilenetv2, tan2019efficientnet, zhang2018shufflenet, xie2017resnext, zhang2022resnest, radosavovic2020regnet, zhang2022parcnet}. 

Therefore, the computer vision community has paid much attention to balancing the effectiveness and complexity of vision transformers. Prior arts concentrate on designing cheaper MHSA as it is the root cause of model complexity. In general, existing works on MHSA complexity alleviation choose to either sparsify the input tokens for vision transformers or reduce dimensions in spatial and channel when computing the attention maps. 
For example, dropping unimportant tokens are proposed in
\cite{rao2021dynamicvit, pan202iared2, meng2022adavit}, hierarchical structure and spatial-reduction attention (SRA) is designed in PVT \cite{wang2021pvt, wang2022pvtv2}, Swin-Transformer \cite{liu2021swin, liu2022swinv2} utilizes shifted window self-attention (SW-SA) to avoid direct global self-attention, and channel-dimension compression when implementing MHSA is used in \cite{yang2022scalablevit}.

These works do reduce the complexity of MHSA effectively, however, none of them asks the following question, ``\textit{Do we really need to obtain every attention map via the expensive correlation computation between the Query and Key signals}?'' Another thing worthy of mentioning is that apart from the MHSA which is widely concerned, the feed-forward network (FFN) originally proposed in the vanilla vision transformer block is usually overlooked by existing research works. Meanwhile, under the circumstance that recent deliberately designed transformer-based backbones, \eg, PVT \cite{wang2021pvt,wang2022pvtv2} and NextViT \cite{li2022nextvit}, have largely reduced complexity compared to the vanilla ViT, we then wonder ``\textit{Is the computation cost of FFN really trivial and what can be further done to balance efficiency and efficacy for FFN}?''

To answer them, we firstly investigate the redundancy of vanilla MHSA module. We visualize attention maps of each block in the three transformer-based backbones DeiT-S\cite{touvron2021deit}, PVTv2\cite{wang2022pvtv2} and NextViT\cite{li2022nextvit}. Taking the $2^{nd}$ block of DeiT-S for example, we shown attention maps in Figure~\ref{fig:intro-q1}. Also, we empirically reveal the mean cos-similarities among attention maps in these backbones exceed 50\%, which will be shown in Table~\ref{tab:ab:hmhsa}. Therefore, the attention maps generated by the complicated Q-K correlation are redundant or correlated.  Then, we analyze model complexity in terms of Params and FLOPs for multiple backbones in Figure~\ref{fig:intro-q2}. It is obvious that FFN consistently contributes a large portion of cost to these models, while it's usually overlooked. Such results are aligned with prior works \cite{Graham2021levit, guo2021cmt}. The above observations suggest that there is still design space for improvement, motivating us to keep in mind the similarity among attention maps for exploiting better MHSA design and to seek for a more well-designed FFN for cost reduction. 

In this work, we therefore propose to design vision transformers with attention map hallucination and FFN compaction. Specifically, largely inspired by \cite{han2020ghostnet} which directly ghosts feature maps from Conv output in CNN, we design a \textit{hallucinated-MHSA (hMHSA)} to hallucinate half of attention maps from the other half using cheaper operations rather than obtaining all attention maps through the expensive Q-K correlation. In this way, the model can effectively alleviate the complexity of MHSA. And surprisingly, these cheaply-generated attention maps are even less redundant, as will be shown in \ref{subsec:redundancy analysis of hMHSA}. Besides, inspired by the fact that redundancy generally exists in FFN \cite{chavan2022slimvit}, we propose our \textit{compact-FFN (cFFN)} for saving FLOPs. Instead of directly reducing the expand ratio (\ie, compressing FFN hidden dimension) \cite{guo2021cmt,Graham2021levit}, which inevitably degrades model capability and performance, we use matrix factorization to lighten the hidden-to-output projection matrix in FFN for redundancy suppression at first, and then off-the-shelf re-parameterization technique \cite{ding2021repvgg} is leveraged to push it being compact and to strengthen its capability. Finally, we apply our hMHSA and cFFN to various backbones, including straight (DeiT \cite{touvron2021deit}), hybrid (NextViT \cite{li2022nextvit}) and hierarchical (PVT \cite{wang2022pvtv2}) structures, experimental results show a 10\%-20\% FLOPs and Params saving can be achieved with competitive performances.
Our contributions are as follows:
\begin{itemize}
\item We exploit correlation/redundancy among attention maps within every MHSA, and propose hallucinated-MHSA (hMHSA), which generates half of attention maps from the other half using cheap operations, to achieve better efficiency-effectiveness balance.
\item 
Compact-FFN (cFFN) is designed, in which hidden-to-output projection matrix is factorized for redundancy reduction and off-the-shelf re-parameterization is utilized for compacting FFN. 
\item Our proposed hMHSA and cFFN are applied to various backbones, we empirically show its effectiveness and achieve a 10\%-20\% FLOPs and Params reduction with competitive performance.
\end{itemize}

\begin{figure*}[t]
\centering
\includegraphics[width=\textwidth]{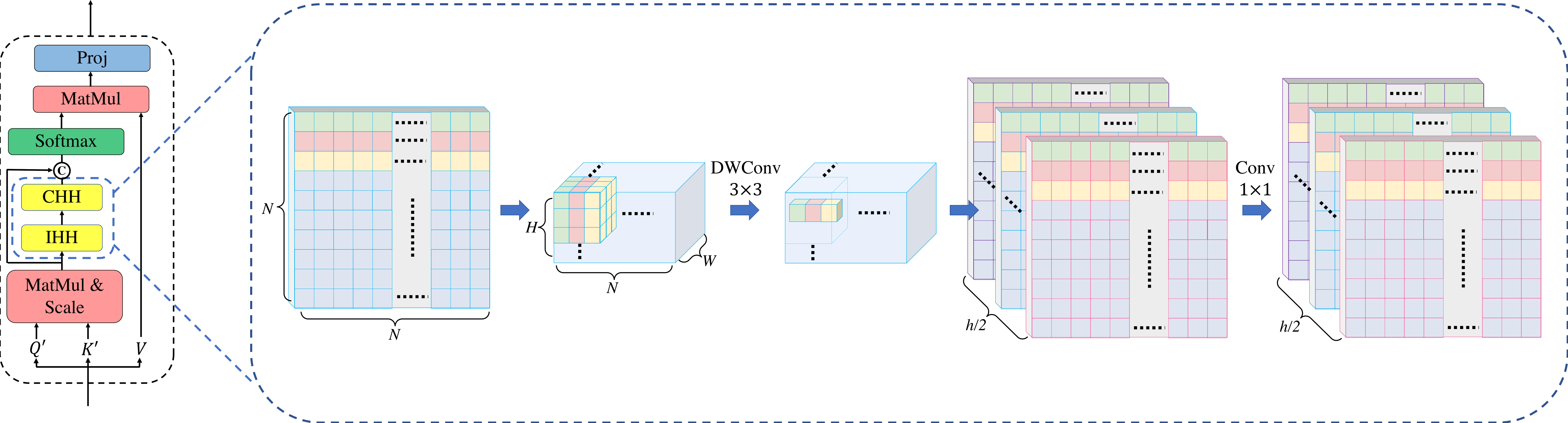}
\caption{The architecture of our proposed hMHSA. Its core idea is to hallucinate half of attention maps based on the other half using cheaper operations of IHH (DWConv $3\times3$) and CHH (Conv $1\times1$). For clarity, we only depict how IHH works on one attention map.} 
\label{fig:arch-hMHSA}
\end{figure*}

\section{Related Work}
\label{sec:related}

\subsection{Token Sparsification}
\label{subsec:Token Sparsification}
The complexity of Vanilla ViT is quadratic with respect to the length of its input tokens, so token sparsification is a natural direction for efficiency improvement. For instance, DynamicViT \cite{rao2021dynamicvit} utilizes a multi-layer perception to predict and drop those less informative tokens. In AdaViT\cite{meng2022adavit}, Adaptive Computation Time(ACT) is propose to discard redundant spatial tokens. EvoViT\cite{xu2022evovit} proposes a slow-fast updating mechanism, which selectively updates informative tokens and less meaningful tokens through different model paths, meanwhile avoiding mistaking token dropping brought by prediction inaccuracy in shadow layers. PatchSlim \cite{tang2022patchslim} adopts the strategy of using effective tokens in the last layer to guide the patch selection in former layers. Meanwhile, DVT\cite{wang2021notallimage16} explores adaptive spatial resolution of patches and invents to automatically configure a proper number of tokens. These solutions all leave the MHSA mechanism and FFN modules unchanged.

\subsection{Dimension Reduction}
\label{subsec:Dimension Reduction}
There are many works focus on reducing the complexity of MHSA, and the common practice is dimension reduction both in spatial or channel axis. 
To reduce the global-attention computational cost, PVT\cite{wang2021pvt, wang2022pvtv2} introduces the spatial-reduction attention(SRA), which reduces the spatial dimension to a constant and therefore relieves complexity. PiT\cite{heo2021pit} uses pooling and depth-wise convolution together to achieve spatial reduction. From the viewpoint of local-attention, Swin-Transformer \cite{liu2021swin} cuts a full image into several windows and performs shifted-window self-attention(SW-SA) mechanism in each window, which also avoids the quadratic complexity. MaxViT \cite{tu2022maxvit} introduces a grid-attention module, which attends cross-window pixels into a sparse, uniform grid overlaid to enhance the global information correlation within local windows. To directly transforms the attention method into linear function, T2T-ViT\cite{yuan2021t2tvit} utilizes the generalized linear attention introduced in Performer\cite{choromanski2020performer}, which also brings linear complexity and improves the efficiency of MHSA. To reduce spatial and channel dimension simultaneous, ScalableViT \cite{yang2022scalablevit} develops the scalable self-attention (SSA), where two scaling factors are introduced to spatial and channel dimensions, respectively. 
However, these aforementioned mechanisms focus on the MHSA and the cost of FFN is generally overlooked in these frameworks. 

\subsection{FFN enhancement}
\label{subsec:FFN enhancement}
The vanilla FFN module in ViTs consists of a linear projection to higher hidden dimension, a non-linear transformation and a linear hidden-to-output projection to recover to the raw dimension. LocalViT\cite{li2021localvit} introduces a $3\times 3$ depth-wise convolution on the hidden dimension, which could enhance the capability of ViTs in local representation extraction, thus offering ViTs the ability to implicitly perceive the relative position information and freeing ViTs from applying complex position embedding. This implementation is also performed in \cite{guo2021cmt, wang2022pvtv2}. LightViT\cite{huang2022lightvit} proposes a bi-dimensional attention module in the FFN to capture the spatial and channel dependencies and refine the features. ParC-Net\cite{zhang2022parcnet} introduces channel-wise attention in the FFN module to conduct data-driven information aggregation. Meanwhile, in order to allocate appropriate computational complexity and design different sizes of models, existing architectures\cite{guo2021cmt, li2022nextvit} tend to scale the expansion ratio of the hidden layer in the FFN module. For instance, LeViT\cite{Graham2021levit} has mentioned FFN might be more expensive, but it simply scales the expansion ratio from 4 to 2, which lacks elaboration and will definitely constrain the capacity of FFN. Our cFFN mainly focus on the compaction of FFN, and it differentiates a lot from these works.

\section{Approach}
\label{sec:Approach}
We propose hMHSA and cFFN for vision transformer construction. Generally, they can be readily used to replace the MHSA and FFN modules of many ViT-based frameworks to achieve better efficacy-efficiency trade-off. 

\subsection{hMHSA Module}
\label{subsec:hMHSA}

Before presenting our proposed hMHSA, we first revisit the vanilla MHSA. For an input $\mathbf{X}\in\mathbb{R}^{N\times C}$ with sequence length of $N$ and channel dimension of $C$, the MHSA module first generates three trainable linear transformation matrices to get query ($\mathbf{Q}$), key ($\mathbf{K}$) and value ($\mathbf{V}$) $\in\mathbb{R}^{N\times C}$. Then, for multi-head calculation, they are reshaped as $\mathbb{R}^{h\times N\times C_h}$, where $h$ stands for the number of heads and $C_h=C / h$ stands for the dimension of each head. Then, the calculation procedure of vanilla MHSA can be summarized as follows: 
\begin{equation} 
\label{eq:MHSA cal}
\begin{split}
& \mathbf{A}_{pre} = {\operatorname{Scale}}(\mathbf{QK}^{\rm T})  \\
& \mathbf{A}_{post} = {\operatorname{Softmax}}(\mathbf{A}_{pre}) \\
& \mathbf{O} = {\operatorname{Proj}}(\mathbf{A}_{post}\mathbf{V})
\end{split}
\end{equation}
where ${\operatorname{Scale}}(\cdot)$ is scaling operation to avoid extreme large result of dot product and ${\operatorname{Proj}}(\cdot)$ is a trainable projection for fusing information from different heads. First, each head $\mathbf{Q_i}$ of $\mathbf{Q}$ conducts scaled dot product with corresponding head $\mathbf{K_i}$ of $\mathbf{K}$, which gives $\mathbf{ A}_{pre}^{i,:,:}\in\mathbb{R}^{N\times N}$. To our convenience, we use $\mathbf{A}_{pre}\in\mathbb{R}^{h\times N\times N}$ to denote the attention maps before ${\operatorname{Softmax}}(\cdot)$, while the attention map after ${\operatorname{Softmax}}(\cdot)$ is denoted as $\mathbf{A}_{post}\in\mathbb{R}^{h\times N\times N}$. Afterwards, $\mathbf{A}_{post}$ serves as the retriever of $\mathbf{V}$, and ${\operatorname{Proj}}(\cdot)$ concatenates all the heads of the retrieval results and applies a fully-connected layer on them to get output $\mathbf{O}$.

Motivated by the discovery of similarities and redundancy within $\mathbf{A}_{pre}$ in the head dimension, we design our hMHSA module to hallucinate half of attention maps from the other half with cheaper operations to reach more efficiency. As shown in Figure~\ref{fig:arch-hMHSA}, the differences between vanilla MHSA module and our hMHSA module are twofold: (1) the generated $\mathbf{\hat{Q}}$ and $\mathbf{\hat{K}}$; (2) the inserted ${\operatorname{IHH}}$ and ${\operatorname{CHH}}$ modules before ${\operatorname{Softmax}}(\cdot)$. To demonstrate clearly, we also formulate the calculation procedure of the hMHSA module as follows: 
\begin{equation} \label{eq:hMHSA cal}
\begin{split}
& \mathbf{A}_{pre}^r = {\operatorname{Scale}}(\mathbf{\hat{Q}}\mathbf{\hat{K}}^T)  \\
& \mathbf{{A}}_{pre}^h = {\operatorname{CHH}}({\operatorname{IHH}}({\mathbf{A}}_{pre}^r))  \\
& \mathbf{{A}}_{pre} = {\operatorname{Concat}}[\mathbf{A}_{pre}^r, \mathbf{A}_{pre}^h] \\
& \mathbf{{A}}_{post} = {\operatorname{Softmax}}(\mathbf{A}_{pre}) \\
& \mathbf{O} = {\operatorname{Proj}}(\mathbf{{A}}_{post}\mathbf{V})
\end{split}
\end{equation}
where ${\operatorname{IHH}}(\cdot)$ stands for Intra-Head Hallucination operation and ${\operatorname{CHH}}(\cdot)$ stands for Cross-Head Hallucination operation. For convenience, we denote the $\textit{real}$ (calculated by $\mathbf{\hat{Q}}$ and $\mathbf{\hat{K}}$) attention maps before ${\operatorname{Softmax}}(\cdot)$ as $\mathbf{A}_{pre}^r$, and the $\textit{hallucinated}$ ones generated by cheaper ${\operatorname{IHH}}(\cdot)$ and ${\operatorname{CHH}}(\cdot)$ as $\mathbf{A}_{pre}^h$.
Different from MHSA, hMHSA generates $\mathbf{\hat{Q}, \hat{K}}\in\mathbb{R}^{N\times \frac{C}{2}}$ and reshape them as $\mathbb{R}^{\frac{h}{2}\times N\times C_h}$ to get $\mathbf{{A}}_{pre}^r \in\mathbb{R}^{\frac{h}{2}\times N \times N}$. Here, for convenience, we suppose the head number $h$ to be even. 
Meanwhile, $\mathbf{V}$ is generated the same as vanilla MHSA module does. 
 
The ${\operatorname{IHH}}(\cdot)$ and ${\operatorname{CHH}}(\cdot)$ operations perform intra-head and cross-head information modeling on the hallucinated $h/2$ attention maps, respectively. They are both implemented using conventional convolutional network modules.
In the ${\operatorname{IHH}}(\cdot)$ phase, each of the real $N\times N$ attention maps in $\mathbf{A}_{pre}^r \in\mathbb{R}^{\frac{h}{2}\times N\times N}$ is reshaped into $N\times H \times W$, where $H$ and $W$ are the original spatial dimensions of input $\mathbf{X}$, changing $\mathbf{{A}}_{pre}^r$ to be $\mathbf{\hat{A}}_{pre}^r \in\mathbb{R}^{N\times \frac{h}{2} \times H \times W}$. Then, a $3\times 3$ depth-wise convolution is applied on the head dimension. The procedure of IHH in \emph{one} attention map is illustrated in the middle part of Figure~\ref{fig:arch-hMHSA}. 

After performing ${\operatorname{IHH}}(\cdot)$, ${\operatorname{CHH}}(\cdot)$ is then applied to model the interaction among different attention heads. Inspired by DeepViT\cite{zhou2021deepvit}, ${\operatorname{CHH}}(\cdot)$ is used to perform  transformation to aggregate the multi-head attention maps along the head dimension. This operation further models the diversity between heads and finally produces $\mathbf{A}_{pre}^h$, and ${\operatorname{CHH}}(\cdot)$ can be conveniently implemented via a $1\times1$ convolution layer (see the right most part of Figure \ref{fig:arch-hMHSA}). 
After generating $\mathbf{A}_{pre}^h$, we concatenate it with $\mathbf{A}_{pre}^r$ along the head dimension to obtain $\mathbf{A}_{pre}$ and the rest of hMHSA is the same as MHSA.

Please note our operation, ${\operatorname{IHH}}(\cdot)$, is designed to be depth-wise convolution, which is also used in GhostNet \cite{han2020ghostnet}. However, they are essentially different. GhostNet adopts depth-wise convolution in the channel dimension mainly considers its lightweight property. In contrast, our operation is conducted in the head dimension, aiming at further modeling affiliation information among neighboring feature points. The element $\mathbf{\hat{A}}_{pre}^r(n,i,h,w)$ represents the correlation between the $n$-th token $P_{1D}$ in the 1D form ($N$) and the $(h,w)$-th token $P_{2D}$ in the 2D form ($H\times W$) of the $i$-th head. As such, the depth-wise $3\times 3$ convolution can learn affinity between different tokens in a local receptive field, \ie, $P_{2D}$ and its 8 neighbours will all contribute to the attention score between $P_{1D}$ and $P_{2D}$. It not only provides stronger modeling ability for MHSA but also saves computational cost.

\subsection{cFFN Module}
\vspace{-5px}
\label{subsec:cFFN}
\begin{figure}[th]
\centering
\includegraphics[width=0.43\textwidth]{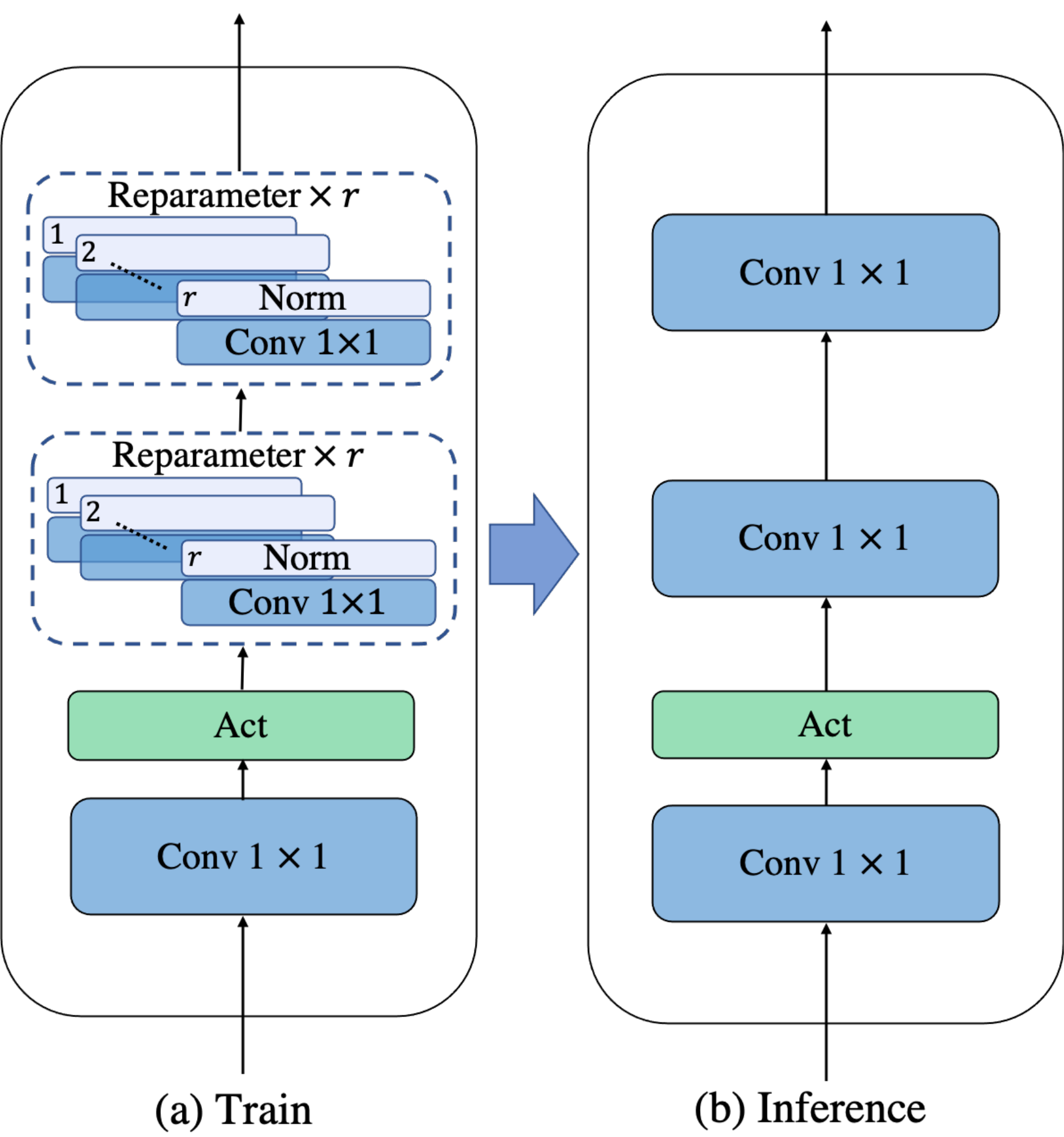}
\caption{The architecture of our proposed cFFN module.}
\label{fig:arch-cFFN}
\end{figure}

First, we also briefly review the vanilla FFN module in a transformer block with a little bit of notation abuse. For a input $\mathbf{X}\in\mathbb{R}^{N\times C}$, when the expand ratio of the FFN hidden layer is $m$, then the Vanilla FFN works as follows:
\begin{equation} \label{eq:vanilla FFN}
\mathbf {O} = ({\operatorname{Act}}(\mathbf{X}\mathbf{M}_1))\mathbf{M}_2,
\end{equation}
where $\mathbf{M}_1\in\mathbb{R}^{C\times mC}$ is the matrix transforming $\mathbf{X}$ to the hidden dimension, ${\operatorname{Act}}(\cdot)$ is a non-linear activation and $\mathbf{M}_2\in\mathbb{R}^{mC\times C}$ is the hidden-to-output projection matrix. 

As demonstrated in MobileNetv2\cite{sandler2018mobilenetv2}, non-linear transformations of low-dimensional manifolds will cause information loss, which behaves as some certain points in the manifold collapse with each other. This means the module will possibly suffer from a decrease in some discriminating points, resulting in inferior modeling ability. Therefore, we declare not to directly decrease the expand ratio for saving computational cost. Instead, we propose the cFFN module to maintain $\mathbf{M}_1$ and the expand ratio to avoid such information loss, while factorizing $\mathbf{M}_2$ for FFN compaction: 
\begin{equation} \label{eq:cFFN}
\mathbf{O} = ({\operatorname{Act}}(\mathbf{X}\mathbf{M}_1))\mathbf{\hat{U_t}}\mathbf{\hat{V_t}}
\end{equation}
where $\mathbf{\hat{U_t}}\in\mathbb{R}^{mC\times k}$, $\mathbf{\hat{V_t}}\in\mathbb{R}^{k\times C}$ represent the factorized two small matrices for $\mathbf{M}_2$ and $k$ is the dimension. Specially, when $k$ equals $mC/(m+1)$, the FLOPs of cFFN is equivalent to that of vanilla FFN. Generally, we set $k=tmC/(m+1), t\in(0, 1)$ and we call $t$ the \textit{compact ratio}. Notably, the factorization could harm the performance, in order for compaction, we leverage re-parameterization(Reparameter) technique \cite{ding2021repvgg} in real implementation. As shown in Figure \ref{fig:arch-cFFN} (a), $r$ branches of Conv $1\times1$ followed by Norm (BN) \cite{ioffe2015batchnorm} are used in training phase for both factorized matrices. In inference phase (Figure \ref{fig:arch-cFFN} (b)), given the linearity of Conv and BN, the $r$ branches are merged into the form of Equation \ref{eq:cFFN}. We would like to highlight our contribution, which is identifying and addressing the long-ignored and poorly handled problem of FFN complexity. With the utilization of the re-parameterization technique as an off-the-shelf method, we were able to compact the FFN module, without sacrificing performance.

\subsection{Complexity Analysis}
\label{Complexity Analysis}

A standard ViT block consists of a MHSA module and a FFN module. In the MHSA module, with a input feature shape of $N\times C$, the three Conv $1\times1$ which transform $\mathbf{X}$ to $\mathbf{Q},\mathbf{K}$ and $\mathbf{V}$ contribute FLOPs of $3NC^2$, the $\mathbf{{Q}}\mathbf{{K}}^{\rm T}$ and $\mathbf{A}_{post}\mathbf{V}$ both contribute $N^2C$, and $\operatorname{Proj}(\cdot)$ contributes $NC^2$. Therefore its complexity in total equals:
\begin{equation}
\label{eqmhsa}
    \mathcal{O}({\operatorname{MHSA}}) = 4NC^2 + 2N^2C
\end{equation}
For the FFN module, suppose the expand ratio to be $m$, then the computational complexity is:
\begin{equation}
\label{eqffn}
    \mathcal{O}({\operatorname{FFN}}) = 2mNC^2
\end{equation}

In our hMHSA module, the complexity of Conv $1\times1$ mapping from $\mathbf{X}$ to $\mathbf{\hat{Q}}$ and $\mathbf{\hat{K}}$ equals $NC^2/2$, respectively. Both mapping from $\mathbf{X}$ to $\mathbf{{V}}$ and the $\operatorname{Proj}(\cdot)$ contribute $NC^2$. The FLOPs of $\mathbf{\hat{Q}}\mathbf{\hat{K}}^{\rm T}$ is $N^2C/2$, while the CHH(DWConv $3\times3$) and IHH(Conv $1\times1$) contributes FLOPs of $9N^2h/2$ and $N^2h^2/4$. Therefore, the total complexity is: 
\begin{equation}
\label{eqhmhsa}
    \mathcal{O}({\operatorname{hMHSA}}) =  3NC^2+3N^2C/2+N^2h^2/4+9N^2h/2
\end{equation}
We can also calculate the complexity of  cFFN, which consists of three mapping whose channel changes from $C$ to $mC$, $mC$ to $k$ and $k$ to $C$ respectively, as:
\begin{equation}
\label{eqcffn}
    \mathcal{O}({\operatorname{cFFN}}) = NmC^2 + NmCk + NkC = (1+t)mNC^2
\end{equation}

From Equation \ref{eqmhsa} and \ref{eqhmhsa}, we can see the FLOPs reduction of our hMHSA compared to vanilla MHSA equals $NC^2+(2C-h^2-18h)N^2/4$. Generally, $C>>h$ therefore the FLOPs reduction of the self-attention module could be ensured. Meanwhile, for the FFN module, our cFFN reduces its FLOPs to a ratio of $(1+t)/2$, with $t<1$, the FLOPs reduction is also non-trivial. 

\section{Experiments}
\label{sec:Experiments}

\subsection{ImageNet Classification}
\label{ImageNet Classification}

\subsubsection{Experiment Settings}
\textbf{Dataset:} ImageNet\cite{deng2009imagenet} is the mainstream classification benchmark in the field of computer vision.  It comprises 1000 classes, including 1.28 million training images and 50,000 validation images. Given its extensive adoption, we employ ImageNet as our dataset for this experiment.

\textbf{Architectures:}
To sufficiently validate the effectiveness and robustness of our proposed hMHSA and cFFN, we choose three recent state-of-the-art vision transformer architectures with different network topology to incorporate our proposed hMHSA and cFFN. Concretely, 
they are the straight ViT architecture DeiT-T and DeiT-S \cite{touvron2021deit}, hierarchical ViT-based architecture PVTv2-b1\cite{wang2022pvtv2} and hybrid convolution and transformer based structures NextViT-S\cite{li2022nextvit}. Please note that these architectures are all with relative small FLOPs and Params, we do not use other heavier backbones, such as ViT-Base and ViT-Large \cite{dosovitskiy2020vit}, to carry out our experiments, because heavier backbones have much more redundancies and we believe empirical results with relatively lightweight architectures are more convincing.    

\begin{table}[ht]
\centering
\resizebox{\columnwidth}{!}
{
    \begin{tabular}{lccc}
    \toprule
    \multirow{2}*{Model} & Params & FLOPs & Acc@1\\
     & (M) & (G) & (\%) \\
    \midrule
    ResNet50\cite{he2016deep}  & 25.6 & 4.1 & 76.2 \\
    ResNet101\cite{he2016deep}  & 44.6 & 7.9 & 80.8 \\
    RegNetY-8G\cite{radosavovic2020regnet}  & 44.2 & 8.0 & 81.7 \\
    ResNeSt50\cite{zhang2022resnest}  & 27.5 & 5.4 & 81.1 \\
    EfficientNet-B3\cite{tan2019efficientnet}  & 12.0 & 1.8 & 81.5 \\
    MobileViTv2-1.0\cite{mehta2022mobilevitv2}  & 4.9 & 4.9 & 78.1 \\
    MobileViTv2-2.0\cite{mehta2022mobilevitv2}  & 18.5 & 7.5 & 81.2 \\
    ConvNeXt-T\cite{liu2022convnext}  & 29.0 & 4.5 & 82.1 \\
    Swin-T\cite{liu2021swin}  & 29.0 & 4.5 & 81.3 \\
    PVT-M\cite{wang2021pvt}  & 44.2 & 6.7 & 81.2 \\
    PVTv2-B2\cite{wang2022pvtv2}  & 25.4 & 4.0 & 82.0 \\
    Twins-SVT-S\cite{chu2021twinsvit}  & 24.0 & 2.9 & 81.7 \\
    PoolFormer-S24\cite{yu2022metaformer}  & 21.1 & 3.4 & 80.3 \\
    PoolFormer-S36\cite{yu2022metaformer}  & 31.2 & 5.0 & 81.4 \\
    CoaT Tiny\cite{xu2021coattu}  & 5.5 & 4.4 & 78.3 \\
    CvT-13\cite{wu2021cvt}  & 20.1 & 4.5 & 81.6 \\
    T2T-ViT-19\cite{yuan2021t2tvit}  & 39.0 & 8.0 & 81.2 \\
    \midrule
    DeiT-T\cite{touvron2021deit}  & 5.72 & 1.26 & 72.2 \\
    DeiT-T\cite{touvron2021deit}\dag  & 4.68/{\color{green}{-18.2\%}} & 1.02/{\color{green}{-19.0\%}} & {\textbf{72.9}} \\
    \midrule
    DeiT-S\cite{touvron2021deit} & 22.05 & 4.60 & 79.9 \\
    DeiT-S\cite{touvron2021deit}\dag  & 17.91/{\color{green}{-18.8\%}} & 3.71/{\color{green}{-19.3\%}} & {\textbf{80.2}} \\
    \midrule
    PVTv2-b1\cite{wang2022pvtv2}  & 14.00 & 2.11 & 78.7 \\
    PVTv2-b1\cite{wang2022pvtv2}\dag  & 12.15/{\color{green}{-13.2\%}} & 1.79/{\color{green}{-15.2\%}} & {\textbf{78.7}} \\
    \midrule
    NextViT-S\cite{li2022nextvit}  & 31.76 & 5.81 & 82.5 \\
    NextViT-S\cite{li2022nextvit}\dag & 27.26/{\color{green}{-13.9\%}} & 5.14/{\color{green}{-11.5\%}} & \textbf{82.5} \\
    \bottomrule
    \end{tabular}
}
\caption{Comparison of different models on ImageNet. \dag \hspace{1px} substitutes modules with our proposed modules.} 
\label{tab:performance}
\end{table}
\textbf{Implementation Details:}
For a fair comparison, we adopt a significant portion of training and data augmentation methods from the official implementations of the three state-of-the-art backbones: DeiT \cite{touvron2021deit}, NextViT \cite{li2022nextvit}, and PVTv2 \cite{wang2022pvtv2}. We make slight modifications to integrate our proposed hMHSA into the original models. 
Concretely, for the three typical state-of-the-art architectures we choose, the number of multi-head-self-attention heads are different from each other and there are odd number of heads even within a single model. In order to satisfy the assumption the head numbers are all even as we assumed in the Approach section (Sec.~\ref{sec:Approach}), before we change MHSA to our hMHSA, we choose to double its head number meanwhile half its channel number for each MHSA module of these backbones to keep the total FLOPs unchanged. In this way, our assumption holds consistently. 
For the cFFN module, we set the compact ratio $t$ to be a constant of $2/3$ to reach a balance between efficiency and performance. Meanwhile, we set the re-parameterization branches number $r$ as $2$. We use AdamW\cite{loshchilov2017adamw} optimizer. Meanwhile, stochastic depth\cite{huang2016stochasticdepth} and repeated augmentation\cite{hoffer2020repeataug1, berman2019repeataug2} are also employed. 

\vspace{-5pt}
\subsubsection{Results and Analysis} 
Table~\ref{tab:performance} presents the ImageNet Classification results, which demonstrate the efficacy of our proposed method. Specifically, we note that PVTv2-b1 exhibits competitive performance with our approach even though it has 13.2\% and 15.2\% fewer parameters and FLOPs as compared to the official implementations. These findings confirm the robustness of our method in handling hierarchical structures and spatial-reduced attention modules. Additionally, NextViT-S equipped with our method achieves comparable accuracy and a reduction of 13.9\% and 11.5\% in Params and FLOPs, respectively. This result highlights the effectiveness of our approach in eliminating redundancies from highly-efficient models. Moreover, DeiT-T and DeiT-S show improvements in their Acc@1 by 0.7\% and 0.3\%, respectively, with the introduction of our proposed modules, demonstrating the effectiveness of our approach in straight structures. Notably, the FLOPs and Params savings of DeiT-T and DeiT-S are also considerable. Thus, our solution reveals that our proposed method enables the improvement of various ViT-based backbones in terms of the efficiency-efficacy trade-off.

\subsection{Ablation Study and Analysis}
\label{Ablation Study}
In order to verify effectiveness of hMHSA and cFFN respectively, in this section, for proof-of-concept purpose, we choose PVTv2-b1\cite{wang2022pvtv2} as our backbone model to exam the design choices of our solution. 
We conduct experiments on a smaller benchmark, the Tiny-ImageNet\cite{le2015tinyimagenet} dataset for saving training cost. Tiny-ImageNet is a subset of ImageNet\cite{deng2009imagenet}, which contains 200 classes within ImageNet. Each class has 500 training images, 50 validation images and 50 test images. 
We modify the last classifier layer's out dimension from 1000 to 200 to match the category numbers of Tiny-ImageNet and the training strategies of PVTv2-b1\cite{wang2022pvtv2} follows its official settings. The only exception is that we train the model at the minimum learning rate for an extra 10 epochs to stabilize the convergence and alleviate the influence of random probabilities. Therefore, by evaluating the model's performance after each of these 10 epochs, we get a more stable and convincing result by averaging the accuracy of these 10 epochs.

\subsubsection{Study on our hMHSA}
\vspace{-10pt}
\begin{table}[hbt]
\centering
\resizebox{\columnwidth}{!}
{
    \begin{tabular}{cccccc}
    \toprule
    Copy & IHH & CHH & Params\textbf{$\downarrow$} & FLOPs\textbf{$\downarrow$} & Acc@1\textbf{$\uparrow$} \\
    \midrule
    \dash & \dash & \dash & 13.6 & 2.11 & 69.3 \\
    \cmark & \dash & \dash & 12.8 & 2.01 & 68.8 \\
    \dash & \cmark & \dash & 12.8 & 2.02 & 69.4 \\
    \dash & \dash & \cmark & 12.8 & 2.02 &  69.2 \\
    \dash & \cmark & \cmark & 12.8 & 2.02 & \textbf{69.7} \\
    \bottomrule
    \end{tabular}
}
\caption{Ablation study of the hMHSA module with different hallucinated head generation methods.}
\label{tab:ab:hMHSA1}
\end{table}

First, we aim to search for an appropriate strategy for generating hallucinated heads from the Q-K generated heads. We make a comparison of different hallucinated head generation methods, including directly copying $\mathbf{A}_{pre}^r$, applying merely $\operatorname{IHH(\cdot)}$ and applying merely $\operatorname{CHH(\cdot)}$. As can been seen from the Table~\ref{tab:ab:hMHSA1}, simply copying harms the performance by 0.5\%. Simply applying $\operatorname{CHH(\cdot)}$ or $\operatorname{IHH(\cdot)}$ reaches comparable performance with the official DeiT-S, while utilizing them simultaneously beats the official accuracy. Therefore, this proves the superiority of our proposed $\operatorname{IHH(\cdot)}$ and $\operatorname{CHH(\cdot)}$. Both interaction among already highly-representative tokens in every head and correlation among different heads could help hMHSA to attain competitive performance as the fully QK-calculated attention maps, whose again verifies that our attention map hallucination strategy is useful.

\begin{table}[hbt]
\centering
\resizebox{\columnwidth}{!}
{
    \begin{tabular}{ccccc}
    \toprule
    IHH & CHH & Params\textbf{$\downarrow$} & FLOPs\textbf{$\downarrow$} & Acc@1\textbf{$\uparrow$} \\
    \midrule
    \dash & \dash & 13.6 & 2.11 & 69.3 \\
    $\mathit{1}$ & $\mathit{2}$ & 12.8 & 2.02 &  \textbf{69.7}\\
    $\mathit{2}$ & $\mathit{1}$ & 12.8 & 2.02 & 69.3\\
    \bottomrule
    \end{tabular}
}
\caption{Ablation study of the operation order of CHH and IHH in our proposed hMHSA Module. $\mathit{1}$ means operating first and $\mathit{2}$ indicates operating afterwards. } 
\label{tab:ab:hMHSA2}
\end{table}

Then, we explore the order of applying both IHH and CHH, \ie, applying $\operatorname{IHH(\cdot)}$ before $\operatorname{CHH(\cdot)}$, and vice versa. The results in Table~\ref{tab:ab:hMHSA2} show that applying $\operatorname{CHH(\cdot)}$ after $\operatorname{IHH(\cdot)}$ beats the other, which demonstrates that modifying intra-head information before fusing cross-head correlation is more reasonable. This is intuitively reasonable, because inter-attention map signals are non-comparable, it is better to model attention relationships in local perceptive field within an attention map, rather than applying $3\times3$ Conv after fusing cross attention map information.

\subsubsection{MHSA v.s. hMHSA}
\label{subsec:redundancy analysis of hMHSA}

\vspace{-10pt}
\begin{table}[!h]
    \centering
    \begin{tabular}{lccc}
    \toprule
    Module Type & DeiT-S & NextViT-S & PVTv2-b1\\
    \midrule
    MHSA & 0.50 & 0.65 & 0.73\\
    hMHSA & 0.41 & 0.60 & 0.70\\
    \bottomrule
    \end{tabular}
    \caption{Quantitative studies of CCS among the attention maps. Calculated by 128 images randomly sampled from the validation set of ImageNet.}
    \label{tab:ab:hmhsa}
\end{table}

In this experiment, we try to empirically measure the redundancy reduction of hMHSA. Motivated by the block-wise similarity calculation proposed in \cite{zhou2021deepvit}, we propose the metric called ``CCS'', short for ``\textit{Contribution Cosine Similarity}'' to measure head-wise similarity. As known in the Attention Mechanism, the rows of the attention map are multiplied with the columns of $\mathbf{V}$ to contribute to the result. Therefore, we quantity the similarity between two different heads base on the similarities between their corresponding rows. First, the similarity between attention heads is calculated as:
\begin{equation}
    S_{n}^{l,m} = 
    \frac{1}{N} 
    \sum_{i=0}^{N-1}
    \frac
    {\langle\mathbf{A}_n^l(i,:),
    \mathbf{A}_n^m(i,:)\rangle}
    {\|\mathbf{A}_n^l(i,:)\|_{2}
    \cdot
    \|\mathbf{A}_n^m(i,:)\|_{2}}
\end{equation}
where $\mathbf{A}_n^l$ denotes the $l^{th}$ head in the $n^{th}$ block and $\langle \cdot, \cdot \rangle$ is the inner product between two vectors. Then, we define the head similarity $S_{n}$ of the $n^{th}$ block as:
\begin{equation}
    S_{n} = 
    \frac{2}{h\times(h-1)} 
    \sum_{l=0}^{N-1}
    \sum_{m=l+1}^{N-1}
    S_{n}^{l,m}
\end{equation}
Finally, we average the overall $B$ block similarity to get the Contribution Cosine Similarity:
\begin{equation}
    {\rm Contribution\ Cosine\ Similarity} = \frac{1}{B}\sum_{n=0}^{B-1}S_n
\end{equation}
Therefore, this metric demonstrates the similarities between different heads directly. As shown in Table~\ref{tab:ab:hmhsa}, we calculate the Contribution Cosine Similarity of the attention maps in the three models. The result shows our hMHSA's values are lower than official ViT block in all the models, which indicates our hMHSA's has less inner similarities among heads in the same block.

\begin{figure}[!t]
\centering
\includegraphics[width=0.4\textwidth]{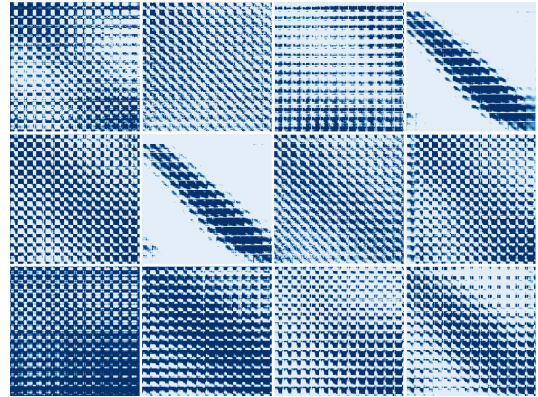}
\caption{Visualization of the 12 maps in the $2^{th}$ block of DeiT-S-ours trained on ImageNet. Left are the 6 vanilla dot-product maps and right are the 6 hallucinated maps.}
\label{fig:visualize_head_ours_block2}
\end{figure}

Moreover, we visualize the attention maps of DeiT-S generated by our hMHSA module in Figure~\ref{fig:visualize_head_ours_block2}. It could be seen that although half of the 12 heads are generate from the other half, actually they have lower similarities among these maps compared to the standard MHSA (Figure~\ref{fig:intro-q1}). For example, only a pair of maps are quite similar, meanwhile others exhibit different attention patterns. This suggest our hMHSA has the ability to flexibly generate less similar attention maps.

\subsubsection{Study on our cFFN}

\begin{table}[h]
\centering
\resizebox{\columnwidth}{!}
{
    \begin{tabular}{lccc}
    \toprule
    Model & Params\textbf{$\downarrow$} & FLOPs\textbf{$\downarrow$} & Acc@1\textbf{$\uparrow$} \\
    \midrule
    PVTv2-b1 & 13.6 & 2.11 &  69.3\\
    \midrule
    reduce $m$, $r=1$ & 12.5 & 1.88 & 68.8 \\
    $t=2/3, r=1$ & 12.5 & 1.88 & 69.1 \\
    $t=2/3, r=2$  & 12.5 & 1.88 & \textbf{70.2} \\
    $t=2/3, r=3$  & 12.5 & 1.88 & 70.0 \\
    \bottomrule
    \end{tabular}
}
\caption{Ablation study of the effectiveness of compaction and the impact of re-parameter branches number $r$ in cFFN.} 
\label{tab:ab:cFFN1}
\end{table}

First, we explore the superiority of FFN compaction over simply reducing $m$ of the hidden layer dimension. For fair comparison, we set compact ratio of cFFN as $t=2/3$ and set m such that vanilla FFN shares the same Params and FLOPs as cFFN. As can be seen in Table~\ref{tab:ab:cFFN1}, simply reducing $m$ severely degrade performance of PVTv2-b1 by 0.5\%. Meanwhile, with our factorization, 69.1\% Acc@1 is achieved, which is 0.3\% higher than directly reducing $m$. It shares that factoring the hidden-to-output mapping matrix is a better design choice. Moreover, we further study the re-parameterization technique\cite{ding2021repvgg} in our cFFN. To better figure out the effect of re-parameterization technique, we ablate over the number $r$ of training-time re-parameterizable branches, specifically 1, 2 and 3. The table shows $r=2$ obtains much better performance than $r=1$, while $r=3$ results in accuracy diminishing. Besides, with more re-parameter branches, training is more costly. Therefore, we choose $r=2$ for the cFFN.  
\begin{table}[h]
\centering
\resizebox{\columnwidth}{!}
{
    \begin{tabular}{ccccc}
    \toprule
    $\mathbf{M}_1$ & $\mathbf{M}_2$ & Params\textbf{$\downarrow$} & FLOPs\textbf{$\downarrow$} & Acc@1\textbf{$\uparrow$} \\
    \midrule
    \dash & \dash & 13.6 & 2.11 & 69.3 \\
    \cmark & \dash & 12.5 & 1.88 &  70.0\\
    \dash & \cmark & 12.5 & 1.88 &  \textbf{70.2}\\
    \bottomrule
    \end{tabular}
}
\caption{Impact of applying factorization and re-parameter on $\mathbf{M}_1$ and $\mathbf{M}_2$, respectively. $r$ equals 2.} 
\label{tab:ab:cFFN2}
\end{table}

\vspace{-8pt}
Then, we show that factorization performed on $\mathbf{M}_2$ is a better choice. We perform factorization and re-parameter on $\mathbf{M}_1$ and $\mathbf{M}_2$, respectively. We can see that compacting $\mathbf{M}_2$ outperforms $\mathbf{M}_1$ by 0.2\% from Table~\ref{tab:ab:cFFN2}. 
The results empirically suggest that compacting $\mathbf{M}_2$ as is done in our implementation is superior. 

\begin{table}[th]
\centering
\resizebox{\columnwidth}{!}
{
    \begin{tabular}{lccc}
    \toprule
    Model & Params\textbf{$\downarrow$} & FLOPs\textbf{$\downarrow$} & Acc@1\textbf{$\uparrow$} \\
    \midrule
    PVTv2-b1 & 13.6 & 2.11 &  69.3\\
    \midrule
    $t=1/2, r=2$ & 12.0 & 1.77 &  69.7 \\
    $t=2/3, r=2$  & 12.5 & 1.88 & \textbf{70.2} \\
    $t=3/4, r=2$  & 12.8 & 1.94 & 70.4 \\
    \bottomrule
    \end{tabular}
}
\caption{Ablation study of different compaction ratio $t$ in the cFFN module. Compacting $\mathbf{M}_2$ is used.} 
\label{tab:ab:cFFN3}
\end{table}

Finally, controlling compact ratio $t$ can help to reach a balance between efficiency and performance. We conduct experiments to empirically seek for appropriate compaction ratio $t$ in our implementation and results are presented in Table~\ref{tab:ab:cFFN3}. We can observe that $t=2/3$ outperforms $t=1/2$ by 0.5\% in Acc@1. When $t$ equals $3/4$, the Params and FLOPs saving becomes quite limited. Therefore, we pick the compact ratio $t$ of $2/3$ as our final choice. 

\subsection{Transfer Study on Downstream Tasks}
To verify the transferability of the proposed method, we show the classification performance on the down-stream tasks, the results are shown in table~\ref{tab:transfer}. In this experiment, we follow common practises to use 384$\times$ 384 and 448$\times$448 for training and testing for all models on CIFAR-100\cite{Krizhevsky2009LearningMLCIFAR100} and Stanford-Dog\cite{StandfordDog}, respectively. The models are initialized from the ImageNet pretrained weights and finetuned on these datasets. From these results, we can find that our variants of the three models are comparable with their original implementation. 

\begin{table}[!h]
    \centering
    \begin{tabular}{lcc}
    \toprule
    Model   &CIFAR-100  &Stanford-Dog\\
    \midrule
         DeiT-S-official / ours    &90.2 / 90.3    &90.0 / 90.5\\
         NextViT-S-official / ours &89.7 / 89.7    &87.9 / 87.8\\
         PVTv2-b1-official  / ours &86.6 / 86.5    &84.9 / 85.1\\
    \bottomrule
    \end{tabular}
    \caption{Performances on downstream datasets. }
    \label{tab:transfer}
\end{table}

\vspace{-15pt}
\section{Limitations and Conclusion}
In this paper, we investigate the similarities (or redundancy) of attention maps in MHSA and propose to leverage such a property via hallucinating half of attention maps from the other half using cheaper operations, dubbed hMHSA. Our hMHSA mechanism discovers a new direction of lightening the cost of MHSA. Besides, we also pay attention to the usually overlooked FFN part of ViT-based architectures and proposed our cFFN strategy to de-redundant it. Empirical results show hMHSA and cFFN can be incorporated into various ViT-based backbones, including straight, hierarchical and hybrid convolution and MHSA structures. These models' complexity in terms of FLOPs and Params can be further reduced, meanwhile, the overall performances of our architectures are quite competitive. However, for proof-of-concept purpose, we only focus on FLOPs and Params for model complexity measurement. No attention is paid to the model throughput currently, because throughput largely depends on hardware acceleration techniques and different implementations could result in dramatic throughput variations. We leave the following as our future work: either hMHSA module design with modern hardware acceleration library (\eg, CUDNN or TensorRT) awareness  or a customized hardware acceleration friendly implementation of our current hMHSA modules. 

\balance
{
\small
\bibliographystyle{ieee_fullname}
\bibliography{egbib}

\begin{thebibliography}{10}\itemsep=-1pt

\bibitem{berman2019repeataug2}
Maxim Berman, Herv{\'e} J{\'e}gou, Andrea Vedaldi, Iasonas Kokkinos, and
  Matthijs Douze.
\newblock Multigrain: a unified image embedding for classes and instances.
\newblock {\em arXiv preprint arXiv:1902.05509}, 2019.

\bibitem{carion2020DETR}
Nicolas Carion, Francisco Massa, Gabriel Synnaeve, Nicolas Usunier, Alexander
  Kirillov, and Sergey Zagoruyko.
\newblock End-to-end object detection with transformers.
\newblock In {\em European conference on computer vision}, pages 213--229.
  Springer, 2020.

\bibitem{chavan2022slimvit}
Arnav Chavan, Zhiqiang Shen, Zhuang Liu, Zechun Liu, Kwang-Ting Cheng, and
  Eric~P Xing.
\newblock Vision transformer slimming: Multi-dimension searching in continuous
  optimization space.
\newblock In {\em Proceedings of the IEEE/CVF Conference on Computer Vision and
  Pattern Recognition}, pages 4931--4941, 2022.

\bibitem{choromanski2020performer}
Krzysztof Choromanski, Valerii Likhosherstov, David Dohan, Xingyou Song,
  Andreea Gane, Tamas Sarlos, Peter Hawkins, Jared Davis, Afroz Mohiuddin,
  Lukasz Kaiser, et~al.
\newblock Rethinking attention with performers.
\newblock {\em arXiv preprint arXiv:2009.14794}, 2020.

\bibitem{chu2021twinsvit}
Xiangxiang Chu, Zhi Tian, Yuqing Wang, Bo Zhang, Haibing Ren, Xiaolin Wei,
  Huaxia Xia, and Chunhua Shen.
\newblock Twins: Revisiting the design of spatial attention in vision
  transformers.
\newblock {\em Advances in Neural Information Processing Systems},
  34:9355--9366, 2021.

\bibitem{deng2009imagenet}
Jia Deng, Wei Dong, Richard Socher, Li-Jia Li, Kai Li, and Li Fei-Fei.
\newblock Imagenet: A large-scale hierarchical image database.
\newblock In {\em 2009 IEEE conference on computer vision and pattern
  recognition}, pages 248--255. Ieee, 2009.

\bibitem{ding2021repvgg}
Xiaohan Ding, Xiangyu Zhang, Ningning Ma, Jungong Han, Guiguang Ding, and Jian
  Sun.
\newblock Repvgg: Making vgg-style convnets great again.
\newblock In {\em Proceedings of the IEEE/CVF Conference on Computer Vision and
  Pattern Recognition}, pages 13733--13742, 2021.

\bibitem{dong2022cswin}
Xiaoyi Dong, Jianmin Bao, Dongdong Chen, Weiming Zhang, Nenghai Yu, Lu Yuan,
  Dong Chen, and Baining Guo.
\newblock Cswin transformer: A general vision transformer backbone with
  cross-shaped windows.
\newblock In {\em Proceedings of the IEEE/CVF Conference on Computer Vision and
  Pattern Recognition}, pages 12124--12134, 2022.

\bibitem{dosovitskiy2020vit}
Alexey Dosovitskiy, Lucas Beyer, Alexander Kolesnikov, Dirk Weissenborn,
  Xiaohua Zhai, Thomas Unterthiner, Mostafa Dehghani, Matthias Minderer, Georg
  Heigold, Sylvain Gelly, et~al.
\newblock An image is worth 16x16 words: Transformers for image recognition at
  scale.
\newblock {\em arXiv preprint arXiv:2010.11929}, 2020.

\bibitem{Graham2021levit}
Benjamin Graham, Alaaeldin El-Nouby, Hugo Touvron, Pierre Stock, Armand Joulin,
  Herv\'e J\'egou, and Matthijs Douze.
\newblock Levit: A vision transformer in convnet's clothing for faster
  inference.
\newblock In {\em Proceedings of the IEEE/CVF International Conference on
  Computer Vision (ICCV)}, pages 12259--12269, October 2021.

\bibitem{guo2021cmt}
Jianyuan Guo, Kai Han, Han Wu, Chang Xu, Yehui Tang, Chunjing Xu, and Yunhe
  Wang.
\newblock Cmt: Convolutional neural networks meet vision transformers.
\newblock {\em arXiv e-prints}, pages arXiv--2107, 2021.

\bibitem{han2020ghostnet}
Kai Han, Yunhe Wang, Qi Tian, Jianyuan Guo, Chunjing Xu, and Chang Xu.
\newblock Ghostnet: More features from cheap operations.
\newblock In {\em Proceedings of the IEEE/CVF conference on computer vision and
  pattern recognition}, pages 1580--1589, 2020.

\bibitem{he2016deep}
Kaiming He, Xiangyu Zhang, Shaoqing Ren, and Jian Sun.
\newblock Deep residual learning for image recognition.
\newblock In {\em Proceedings of the IEEE conference on computer vision and
  pattern recognition}, pages 770--778, 2016.

\bibitem{heo2021pit}
Byeongho Heo, Sangdoo Yun, Dongyoon Han, Sanghyuk Chun, Junsuk Choe, and
  Seong~Joon Oh.
\newblock Rethinking spatial dimensions of vision transformers.
\newblock In {\em Proceedings of the IEEE/CVF International Conference on
  Computer Vision}, pages 11936--11945, 2021.

\bibitem{hoffer2020repeataug1}
Elad Hoffer, Tal Ben-Nun, Itay Hubara, Niv Giladi, Torsten Hoefler, and Daniel
  Soudry.
\newblock Augment your batch: Improving generalization through instance
  repetition.
\newblock In {\em Proceedings of the IEEE/CVF Conference on Computer Vision and
  Pattern Recognition}, pages 8129--8138, 2020.

\bibitem{huang2016stochasticdepth}
Gao Huang, Yu Sun, Zhuang Liu, Daniel Sedra, and Kilian~Q Weinberger.
\newblock Deep networks with stochastic depth.
\newblock In {\em European conference on computer vision}, pages 646--661.
  Springer, 2016.

\bibitem{huang2022lightvit}
Tao Huang, Lang Huang, Shan You, Fei Wang, Chen Qian, and Chang Xu.
\newblock Lightvit: Towards light-weight convolution-free vision transformers.
\newblock {\em arXiv preprint arXiv:2207.05557}, 2022.

\bibitem{ioffe2015batchnorm}
Sergey Ioffe and Christian Szegedy.
\newblock Batch normalization: Accelerating deep network training by reducing
  internal covariate shift.
\newblock In {\em International conference on machine learning}, pages
  448--456. PMLR, 2015.

\bibitem{StandfordDog}
Aditya Khosla, Nityananda Jayadevaprakash, Bangpeng Yao, and Li Fei-Fei.
\newblock Novel dataset for fine-grained image categorization.
\newblock In {\em First Workshop on Fine-Grained Visual Categorization, IEEE
  Conference on Computer Vision and Pattern Recognition}, Colorado Springs, CO,
  June 2011.

\bibitem{Krizhevsky2009LearningMLCIFAR100}
Alex Krizhevsky.
\newblock Learning multiple layers of features from tiny images.
\newblock 2009.

\bibitem{le2015tinyimagenet}
Ya Le and Xuan Yang.
\newblock Tiny imagenet visual recognition challenge.
\newblock {\em CS 231N}, 7(7):3, 2015.

\bibitem{li2022nextvit}
Jiashi Li, Xin Xia, Wei Li, Huixia Li, Xing Wang, Xuefeng Xiao, Rui Wang, Min
  Zheng, and Xin Pan.
\newblock Next-vit: Next generation vision transformer for efficient deployment
  in realistic industrial scenarios.
\newblock {\em arXiv preprint arXiv:2207.05501}, 2022.

\bibitem{li2022vitdet}
Yanghao Li, Hanzi Mao, Ross Girshick, and Kaiming He.
\newblock Exploring plain vision transformer backbones for object detection.
\newblock {\em arXiv preprint arXiv:2203.16527}, 2022.

\bibitem{li2021localvit}
Yawei Li, Kai Zhang, Jiezhang Cao, Radu Timofte, and Luc Van~Gool.
\newblock Localvit: Bringing locality to vision transformers.
\newblock {\em arXiv preprint arXiv:2104.05707}, 2021.

\bibitem{liu2022swinv2}
Ze Liu, Han Hu, Yutong Lin, Zhuliang Yao, Zhenda Xie, Yixuan Wei, Jia Ning, Yue
  Cao, Zheng Zhang, Li Dong, et~al.
\newblock Swin transformer v2: Scaling up capacity and resolution.
\newblock In {\em Proceedings of the IEEE/CVF Conference on Computer Vision and
  Pattern Recognition}, pages 12009--12019, 2022.

\bibitem{liu2021swin}
Ze Liu, Yutong Lin, Yue Cao, Han Hu, Yixuan Wei, Zheng Zhang, Stephen Lin, and
  Baining Guo.
\newblock Swin transformer: Hierarchical vision transformer using shifted
  windows.
\newblock In {\em Proceedings of the IEEE/CVF International Conference on
  Computer Vision}, pages 10012--10022, 2021.

\bibitem{liu2022convnext}
Zhuang Liu, Hanzi Mao, Chao-Yuan Wu, Christoph Feichtenhofer, Trevor Darrell,
  and Saining Xie.
\newblock A convnet for the 2020s.
\newblock In {\em Proceedings of the IEEE/CVF Conference on Computer Vision and
  Pattern Recognition}, pages 11976--11986, 2022.

\bibitem{loshchilov2017adamw}
Ilya Loshchilov and Frank Hutter.
\newblock Decoupled weight decay regularization.
\newblock {\em arXiv preprint arXiv:1711.05101}, 2017.

\bibitem{mehta2022mobilevitv2}
Sachin Mehta and Mohammad Rastegari.
\newblock Separable self-attention for mobile vision transformers.
\newblock {\em arXiv preprint arXiv:2206.02680}, 2022.

\bibitem{meng2022adavit}
Lingchen Meng, Hengduo Li, Bor-Chun Chen, Shiyi Lan, Zuxuan Wu, Yu-Gang Jiang,
  and Ser-Nam Lim.
\newblock Adavit: Adaptive vision transformers for efficient image recognition.
\newblock In {\em Proceedings of the IEEE/CVF Conference on Computer Vision and
  Pattern Recognition}, pages 12309--12318, 2022.

\bibitem{pan202iared2}
Bowen Pan, Rameswar Panda, Yifan Jiang, Zhangyang Wang, Rogerio Feris, and Aude
  Oliva.
\newblock Ia-red$^2$: Interpretability-aware redundancy reduction for vision
  transformers.
\newblock {\em Advances in Neural Information Processing Systems},
  34:24898--24911, 2021.

\bibitem{radosavovic2020regnet}
Ilija Radosavovic, Raj~Prateek Kosaraju, Ross Girshick, Kaiming He, and Piotr
  Doll{\'a}r.
\newblock Designing network design spaces.
\newblock In {\em Proceedings of the IEEE/CVF conference on computer vision and
  pattern recognition}, pages 10428--10436, 2020.

\bibitem{ranftl2021vitdensepred}
Ren{\'e} Ranftl, Alexey Bochkovskiy, and Vladlen Koltun.
\newblock Vision transformers for dense prediction.
\newblock In {\em Proceedings of the IEEE/CVF International Conference on
  Computer Vision}, pages 12179--12188, 2021.

\bibitem{rao2021dynamicvit}
Yongming Rao, Wenliang Zhao, Benlin Liu, Jiwen Lu, Jie Zhou, and Cho-Jui Hsieh.
\newblock Dynamicvit: Efficient vision transformers with dynamic token
  sparsification.
\newblock {\em Advances in neural information processing systems},
  34:13937--13949, 2021.

\bibitem{sandler2018mobilenetv2}
Mark Sandler, Andrew Howard, Menglong Zhu, Andrey Zhmoginov, and Liang-Chieh
  Chen.
\newblock Mobilenetv2: Inverted residuals and linear bottlenecks.
\newblock In {\em Proceedings of the IEEE conference on computer vision and
  pattern recognition}, pages 4510--4520, 2018.

\bibitem{simonyan2014vgg}
Karen Simonyan and Andrew Zisserman.
\newblock Very deep convolutional networks for large-scale image recognition.
\newblock {\em arXiv preprint arXiv:1409.1556}, 2014.

\bibitem{strudel2021segmenter}
Robin Strudel, Ricardo Garcia, Ivan Laptev, and Cordelia Schmid.
\newblock Segmenter: Transformer for semantic segmentation.
\newblock In {\em Proceedings of the IEEE/CVF International Conference on
  Computer Vision}, pages 7262--7272, 2021.

\bibitem{tan2019efficientnet}
Mingxing Tan and Quoc Le.
\newblock Efficientnet: Rethinking model scaling for convolutional neural
  networks.
\newblock In {\em International conference on machine learning}, pages
  6105--6114. PMLR, 2019.

\bibitem{tang2022patchslim}
Yehui Tang, Kai Han, Yunhe Wang, Chang Xu, Jianyuan Guo, Chao Xu, and Dacheng
  Tao.
\newblock Patch slimming for efficient vision transformers.
\newblock In {\em Proceedings of the IEEE/CVF Conference on Computer Vision and
  Pattern Recognition}, pages 12165--12174, 2022.

\bibitem{touvron2021deit}
Hugo Touvron, Matthieu Cord, Matthijs Douze, Francisco Massa, Alexandre
  Sablayrolles, and Herv{\'e} J{\'e}gou.
\newblock Training data-efficient image transformers \& distillation through
  attention.
\newblock In {\em International Conference on Machine Learning}, pages
  10347--10357. PMLR, 2021.

\bibitem{tu2022maxvit}
Zhengzhong Tu, Hossein Talebi, Han Zhang, Feng Yang, Peyman Milanfar, Alan
  Bovik, and Yinxiao Li.
\newblock Maxvit: Multi-axis vision transformer.
\newblock {\em arXiv preprint arXiv:2204.01697}, 2022.

\bibitem{wang2021pvt}
Wenhai Wang, Enze Xie, Xiang Li, Deng-Ping Fan, Kaitao Song, Ding Liang, Tong
  Lu, Ping Luo, and Ling Shao.
\newblock Pyramid vision transformer: A versatile backbone for dense prediction
  without convolutions.
\newblock In {\em Proceedings of the IEEE/CVF International Conference on
  Computer Vision}, pages 568--578, 2021.

\bibitem{wang2022pvtv2}
Wenhai Wang, Enze Xie, Xiang Li, Deng-Ping Fan, Kaitao Song, Ding Liang, Tong
  Lu, Ping Luo, and Ling Shao.
\newblock Pvt v2: Improved baselines with pyramid vision transformer.
\newblock {\em Computational Visual Media}, 8(3):415--424, 2022.

\bibitem{wang2021notallimage16}
Yulin Wang, Rui Huang, Shiji Song, Zeyi Huang, and Gao Huang.
\newblock Not all images are worth 16x16 words: Dynamic transformers for
  efficient image recognition.
\newblock {\em Advances in Neural Information Processing Systems},
  34:11960--11973, 2021.

\bibitem{wu2021cvt}
Haiping Wu, Bin Xiao, Noel Codella, Mengchen Liu, Xiyang Dai, Lu Yuan, and Lei
  Zhang.
\newblock Cvt: Introducing convolutions to vision transformers.
\newblock In {\em Proceedings of the IEEE/CVF International Conference on
  Computer Vision}, pages 22--31, 2021.

\bibitem{xie2017resnext}
Saining Xie, Ross Girshick, Piotr Doll{\'a}r, Zhuowen Tu, and Kaiming He.
\newblock Aggregated residual transformations for deep neural networks.
\newblock In {\em Proceedings of the IEEE conference on computer vision and
  pattern recognition}, pages 1492--1500, 2017.

\bibitem{xu2021coattu}
Weijian Xu, Yifan Xu, Tyler Chang, and Zhuowen Tu.
\newblock Co-scale conv-attentional image transformers.
\newblock In {\em Proceedings of the IEEE/CVF International Conference on
  Computer Vision}, pages 9981--9990, 2021.

\bibitem{xu2022evovit}
Yifan Xu, Zhijie Zhang, Mengdan Zhang, Kekai Sheng, Ke Li, Weiming Dong, Liqing
  Zhang, Changsheng Xu, and Xing Sun.
\newblock Evo-vit: Slow-fast token evolution for dynamic vision transformer.
\newblock In {\em Proceedings of the AAAI Conference on Artificial
  Intelligence}, volume~36, pages 2964--2972, 2022.

\bibitem{yang2022scalablevit}
Rui Yang, Hailong Ma, Jie Wu, Yansong Tang, Xuefeng Xiao, Min Zheng, and Xiu
  Li.
\newblock Scalablevit: Rethinking the context-oriented generalization of vision
  transformer.
\newblock {\em arXiv preprint arXiv:2203.10790}, 2022.

\bibitem{yu2022metaformer}
Weihao Yu, Mi Luo, Pan Zhou, Chenyang Si, Yichen Zhou, Xinchao Wang, Jiashi
  Feng, and Shuicheng Yan.
\newblock Metaformer is actually what you need for vision.
\newblock In {\em Proceedings of the IEEE/CVF Conference on Computer Vision and
  Pattern Recognition}, pages 10819--10829, 2022.

\bibitem{yuan2021t2tvit}
Li Yuan, Yunpeng Chen, Tao Wang, Weihao Yu, Yujun Shi, Zi-Hang Jiang,
  Francis~EH Tay, Jiashi Feng, and Shuicheng Yan.
\newblock Tokens-to-token vit: Training vision transformers from scratch on
  imagenet.
\newblock In {\em Proceedings of the IEEE/CVF International Conference on
  Computer Vision}, pages 558--567, 2021.

\bibitem{zhang2022parcnet}
Haokui Zhang, Wenze Hu, and Xiaoyu Wang.
\newblock Parc-net: Position aware circular convolution with merits from
  convnets and transformer.
\newblock {\em networks (ConvNets)}, 5(33):21, 2022.

\bibitem{zhang2022dino}
Hao Zhang, Feng Li, Shilong Liu, Lei Zhang, Hang Su, Jun Zhu, Lionel~M Ni, and
  Heung-Yeung Shum.
\newblock Dino: Detr with improved denoising anchor boxes for end-to-end object
  detection.
\newblock {\em arXiv preprint arXiv:2203.03605}, 2022.

\bibitem{zhang2022resnest}
Hang Zhang, Chongruo Wu, Zhongyue Zhang, Yi Zhu, Haibin Lin, Zhi Zhang, Yue
  Sun, Tong He, Jonas Mueller, R Manmatha, et~al.
\newblock Resnest: Split-attention networks.
\newblock In {\em Proceedings of the IEEE/CVF Conference on Computer Vision and
  Pattern Recognition}, pages 2736--2746, 2022.

\bibitem{zhang2018shufflenet}
Xiangyu Zhang, Xinyu Zhou, Mengxiao Lin, and Jian Sun.
\newblock Shufflenet: An extremely efficient convolutional neural network for
  mobile devices.
\newblock In {\em Proceedings of the IEEE conference on computer vision and
  pattern recognition}, pages 6848--6856, 2018.

\bibitem{zheng2021SETR}
Sixiao Zheng, Jiachen Lu, Hengshuang Zhao, Xiatian Zhu, Zekun Luo, Yabiao Wang,
  Yanwei Fu, Jianfeng Feng, Tao Xiang, Philip~HS Torr, et~al.
\newblock Rethinking semantic segmentation from a sequence-to-sequence
  perspective with transformers.
\newblock In {\em Proceedings of the IEEE/CVF conference on computer vision and
  pattern recognition}, pages 6881--6890, 2021.

\bibitem{zhou2021deepvit}
Daquan Zhou, Bingyi Kang, Xiaojie Jin, Linjie Yang, Xiaochen Lian, Zihang
  Jiang, Qibin Hou, and Jiashi Feng.
\newblock Deepvit: Towards deeper vision transformer.
\newblock {\em arXiv preprint arXiv:2103.11886}, 2021.

\bibitem{zhu2020deformabledetr}
Xizhou Zhu, Weijie Su, Lewei Lu, Bin Li, Xiaogang Wang, and Jifeng Dai.
\newblock Deformable detr: Deformable transformers for end-to-end object
  detection.
\newblock {\em arXiv preprint arXiv:2010.04159}, 2020.

\end{thebibliography}
}

\end{document}